\pdfoutput=1

\documentclass[11pt]{article}

\usepackage[review]{acl}
\usepackage{graphicx}
\usepackage{booktabs}
\usepackage{subcaption}

\usepackage{times}
\usepackage{latexsym}
\usepackage[T1]{fontenc}

\usepackage[utf8]{inputenc}

\usepackage{microtype}

\title{Construction of English Resume Corpus and Test with Pre-trained Language Models}

\author{Chengguang Gan \And Tatsunori Mori \\
\AND
  Yokohama National University, Japan \\
  \texttt{gan-chengguang-pw@ynu.jp, tmori@ynu.ac.jp} \\}

\begin{document}
\maketitle
\begin{abstract}
Information extraction(IE) has always been one of the essential tasks of NLP. Moreover, one of the most critical application scenarios of information extraction is the information extraction of resumes. Constructed text is obtained by classifying each part of the resume. It is convenient to store these texts for later search and analysis. Furthermore, the constructed resume data can also be used in the AI resume screening system. Significantly reduce the labor cost of HR. This study aims to transform the information extraction task of resumes into a simple sentence classification task. Based on the English resume dataset produced by the prior study. The classification rules are improved to create a larger and more fine-grained classification dataset of resumes. This corpus is also used to test some current mainstream Pre-training language models (PLMs) performance.Furthermore, in order to explore the relationship between the number of training samples and the correctness rate of the resume dataset, we also performed comparison experiments with training sets of different train set sizes. The final multiple experimental results show that the resume dataset with improved annotation rules and increased sample size of the dataset improves the accuracy of the original resume dataset.
\end{abstract}

\section{Introduction}
As artificial intelligence develops, using artificial intelligence instead of HR for resume screening has always been the focus of research. And the accuracy of resume screening depends on the precision of resume information extraction.Hence, it is crucial to improve the precision of resume extraction for the subsequent steps of various analyses performance of resumes. The previous study on resume information extraction tends to use the Bi-LSTM-CRF model for 
Name Entity Recognition(NER) of resume text\cite{2018Entity}.Although this method extracts the 
resume information (e.g. Personal information, Name, Address, Gender, Birth) with high accuracy, 
it also loses some original verbal expression information. For example, the description of one's 
future career goals, requires complete sentences that cannot be extracted by the NER method. As an 
AI system that scores the candidate's resume, the career object is also part of the score. In 
summary, sentences such as these should not be ignored. Hence, in the prior study, the task of 
resume information extraction is transformed into a sentence classification 
task\cite{weko_213712_1}
. Firstly, the various resume formats were 
converted into a uniform txt document. Then the sentences were classified after dividing them by 
sentence units. The classified sentences are used in the subsequent AI scoring system for 
resumes. The pilot study segmented and annotated 500 of the 15,000 original CVs from 
Kaggle.\footnote{\raggedright\url{https://www.kaggle.com/datasets/oo7kartik/resume-text-batch}}
Five categories of tags were set: \emph{experience}, \emph{knowledge}, \emph{education}, \emph{project} and 
\emph{others}\footnote{\raggedright\url{https://www.kaggle.com/datasets/chingkuangkam/resume-text-classification-dataset}}. The pilot study annotated resume dataset has problems, such as unclear 
classification label boundaries and fewer categories. 
Also, a dataset of 500 resumes with a total of 40,000 sentences in the tagging is sufficient for 
PLMs to fine-tune. If the dataset sample is increased, can the model's performance continue to 
improve.
\par To resolve all these problems, we improved the classification labels of resumes and used them to label a new resume classification dataset. To find out how many training samples can satisfy the fine-tune requirement of PLMs, we annotated 1000 resumes with a total of 78000 sentences. Furthermore, various experiments have been performed on the newly created resume dataset using the current mainstream PLMs.

\section{Related Work}

Since the last century, resume information extraction has been a critical applied research subfield in IE. In earlier studies, methods such as rule-based and dictionary matching were used to extract specific information from resumes\cite{mooney1999relational}. HMM, and SVM methods extract information such as a person's name and phone number from resume information\cite{yu2005resume}.Related Resume Corpus Construct study has an extensive resume corpus in Chinese\cite{su2019resume}.

\section{Corpus Construction}

\subsection{Annotation Rule}

\begin{figure}[t]
\centering
\includegraphics[width=219 pt]{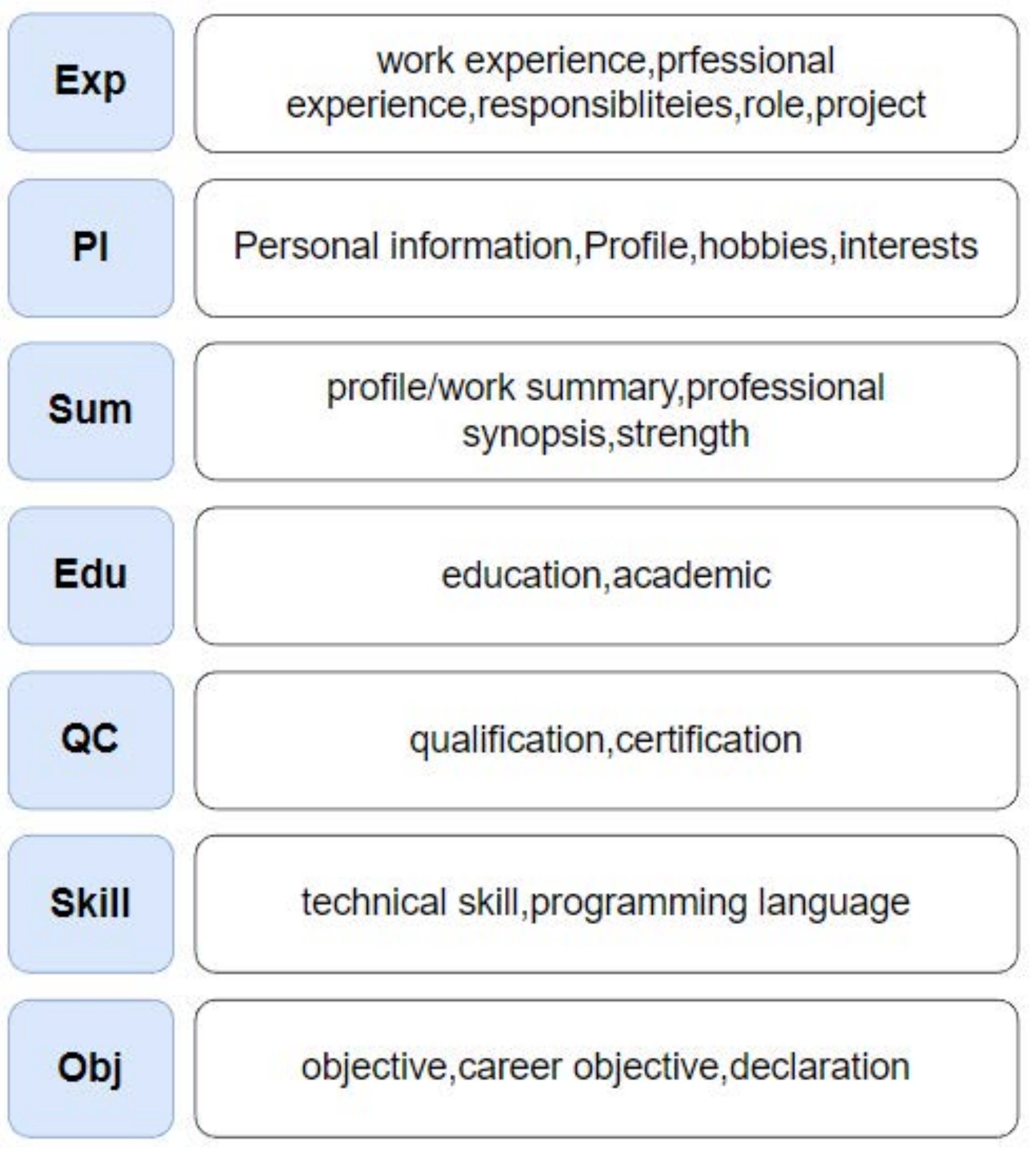}
\caption{Resume annotation rules diagram.}
\label{figure1}
\end{figure}

We increased the number of categories from 5 to 7 in order to discriminate the various parts of the resume more carefully.As shown in Figure \ref{figure1}, the blue block on the left is the abbreviation of the seven classification labels, and on the right is the name of the resume section corresponding to the label. The full names of the seven labels are \emph{Experience}, \emph{Personal Information}, \emph{Summary}, \emph{Education}, \emph{Qualifications}, \emph{Skill}, and \emph{Object}. The newly developed classification rules make it possible to have a clear attribution for each item in the resume. It will not cause the neglect of some sentences in the resume, as there are \emph{other} labels in the prior study.

\subsection{Annotation Tool}

\begin{figure}[t]
\centering
\includegraphics[width=219 pt]{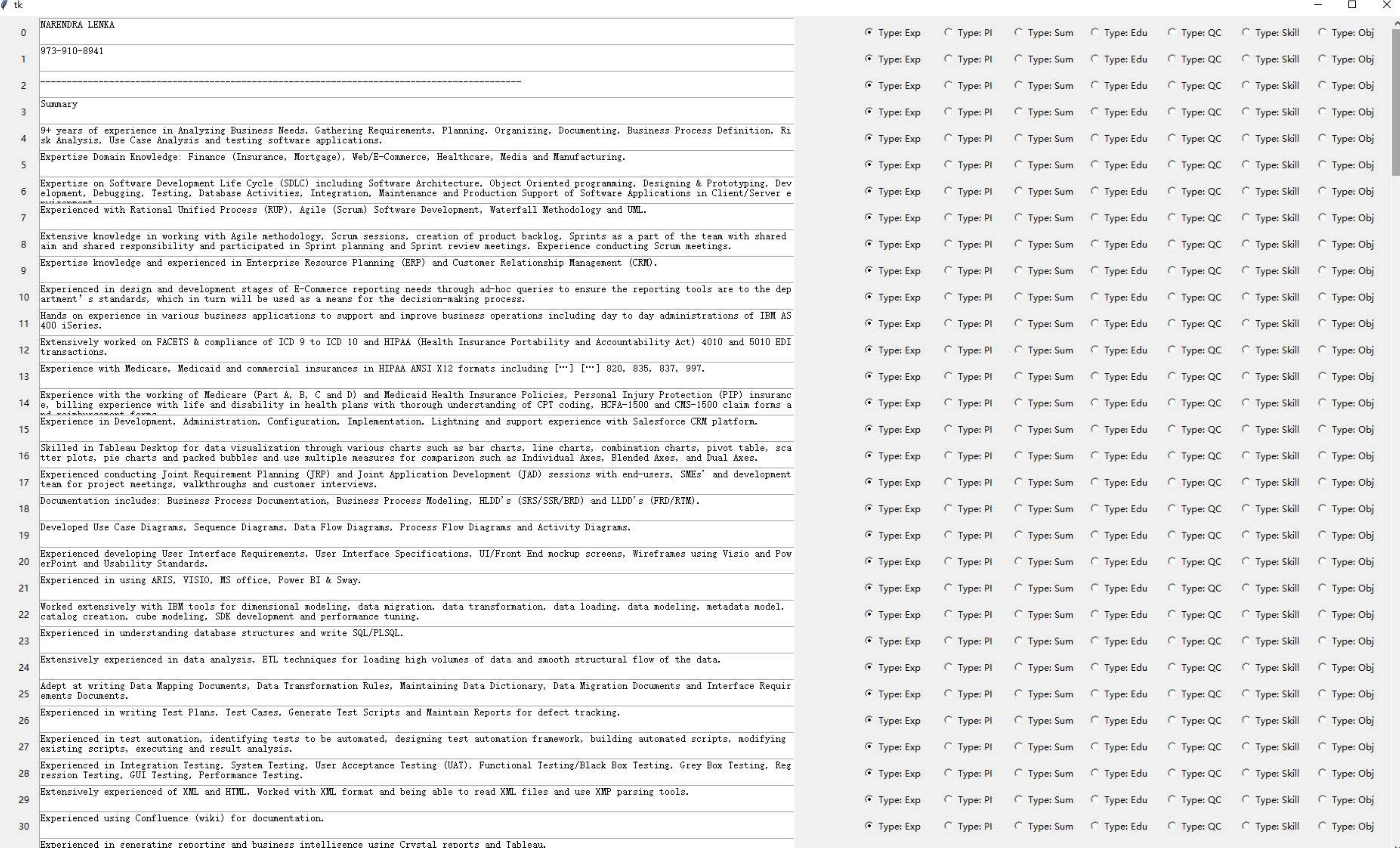}
\caption{The operation interface of the resume annotation tool.}
\label{figure2}
\end{figure}

In order to label resume datasets faster and more accurately, we developed a simple annotation program based on Tkinter\footnote{https://docs.python.org/ja/3/library/tkinter.html}. The operation interface of the resume annotation tool. This tool automatically recognizes original resumes in pdf, Docx, and txt formats. It can also segment all the sentences in the original resume according to a simple rule-based approach.Figure \ref{figure2} shows the sample interface of the resume annotation tool. On the left are the sentences split by rule-based, and on the right are seven buttons that can be selected individually. After the sentence annotation of a whole resume is completed, a separate txt file will be automatically exported after closing the window, and the sentence annotation window for the next resume will be started automatically. Examples of annotated resume 
sentences can be seen in the Appendix \ref{sec:appendix}.

\begin{table*}[!t]
\centering
\begin{tabular}{lccccccccccccc}
\toprule[2pt]
Sample    &10000 &15000 &20000 &25000 &30000 &35000 &40000 &45000 &50000 &55000 \\
\midrule
Valid  &83 &83.7 &84.9 &84.9 &85.6 &86 &86.1 &\textbf{86.6} & 85.9&85.9 \\
Test    &83.5 &84.3 &85.3 &85.6 &84.6 &85.4 &\textbf{85.9} &\textbf{85.9} &85.8 &85.1\\
\bottomrule[2pt]
\end{tabular}
\caption{\label{table1}
The first row indicates the number of training sets. The following two rows indicate the F1-score of the validation set and test set corresponding to the number of training samples.}
\end{table*}

\section{Experiments Set}

In this section, we will perform various test experiments on the new-constructed resume dataset. First, we compared the performance of the BERT\cite{devlin2018bert} model on the original resume corpus and the newly constructed resume dataset. Furthermore, four mainstream PLMs models are selected to test the resume dataset performance: BERT,ALBERT\cite{lan2019albert},RoBERTa\cite{liu2019roberta}, and T5\cite{raffel2020exploring}. For the fairness of the experiment, the size with the most similar parameters was chosen for each of the four models(BERT$_{\mbox{\scriptsize large}}$, ALBERT$_{\mbox{\scriptsize xxlarge}}$, RoBERTa$_{\mbox{\scriptsize large}}$, T5$_{\mbox{\scriptsize large}}$). The evaluation metrics for all experiments were F1-micro. The training set, validation set, and test set are randomly divided in the ratio of 7:1.5:1.5.And each experiment was performed three times to take the average of the results.

\begin{table}[!ht]
\centering
\begin{tabular}{lr}
\toprule[2pt]
\makebox[0.2\textwidth][l]{\textbf{Model}} & \makebox[0.2\textwidth][r]{\textbf{F1-score}}\\
\midrule
BERT*$_{\mbox{\scriptsize large}}$(baseline)  & 85.97\\
BERT$_{\mbox{\scriptsize large}}$ & \textbf{86.67}\\
ALBERT$_{\mbox{\scriptsize large}}$ & \textbf{86.40}\\
RoBERTa$_{\mbox{\scriptsize large}}$ &\textbf{87.00}\\
T5$_{\mbox{\scriptsize large}}$ & \textbf{87.35}\\
\bottomrule[2pt]
\end{tabular}
\caption{\label{table2}
The first column * show accuracy of resume dataset before improvement.}
\end{table}

\section{Result}

\subsection{Pre-train Models Test}

As shown in Table \ref{table2}, the new resume corpus ameliorated by 0.70$\%$ over the original 
dataset F1-score for the same BERT model. RoBERTa and T5 scores improved by 1.03$\%$ and 1.38$\%$ over 
baseline, respectively. The above results are also consistent with the ranking of the four PLMs in 
terms of their performance in various benchmark tests of NLP.

\subsection{Sample Size Affects Experiment}

\begin{figure}[t]
\centering
\includegraphics[width=219 pt]{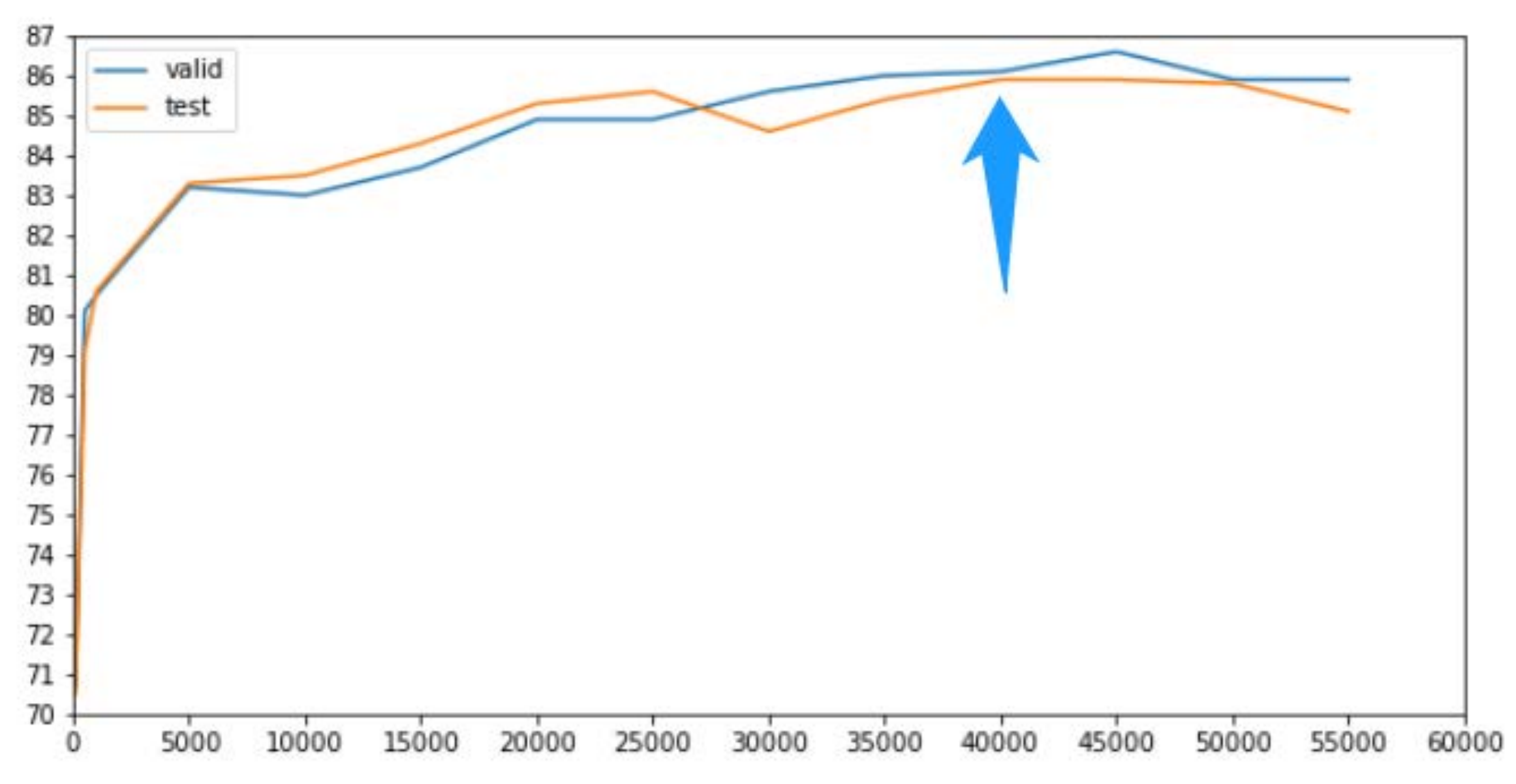}
\caption{F1-score of different training samples.}
\label{figure3}
\end{figure}

In order to find out how many samples can bring out the maximum performance of the model, we divide the data set into training set 58000: validation set 10000: test set 10000. As shown in Table \ref{table1}, the scores of the validation and test sets for different sample sizes. The model scores are tested from the 58000 training set, starting from 5000 and increasing the number of training samples every 5000. The highest score in the validation set is 86.6 when the training sample equals 45000. the highest score in the test set is 85.9 when the training sample equals 40000 and 45000.In order to visualize the relationship between the number of training samples and performance, we plotted the graphs (As Figure \ref{figure3}).It can be seen that as the number of training samples increases, the correctness of the model rises. Finally, the model's performance reaches the highest point when the training samples are increased to 40,000. From the experimental results, for the PLMs, this resume corpus above 40,000 is sufficient for the model's maximum performance. The results also prove that the new resume corpus, which doubles the sample size, is significant compared to the original resum corpus.

\section{Analysis}

\begin{figure}[t]
\centering
\includegraphics[width=219 pt]{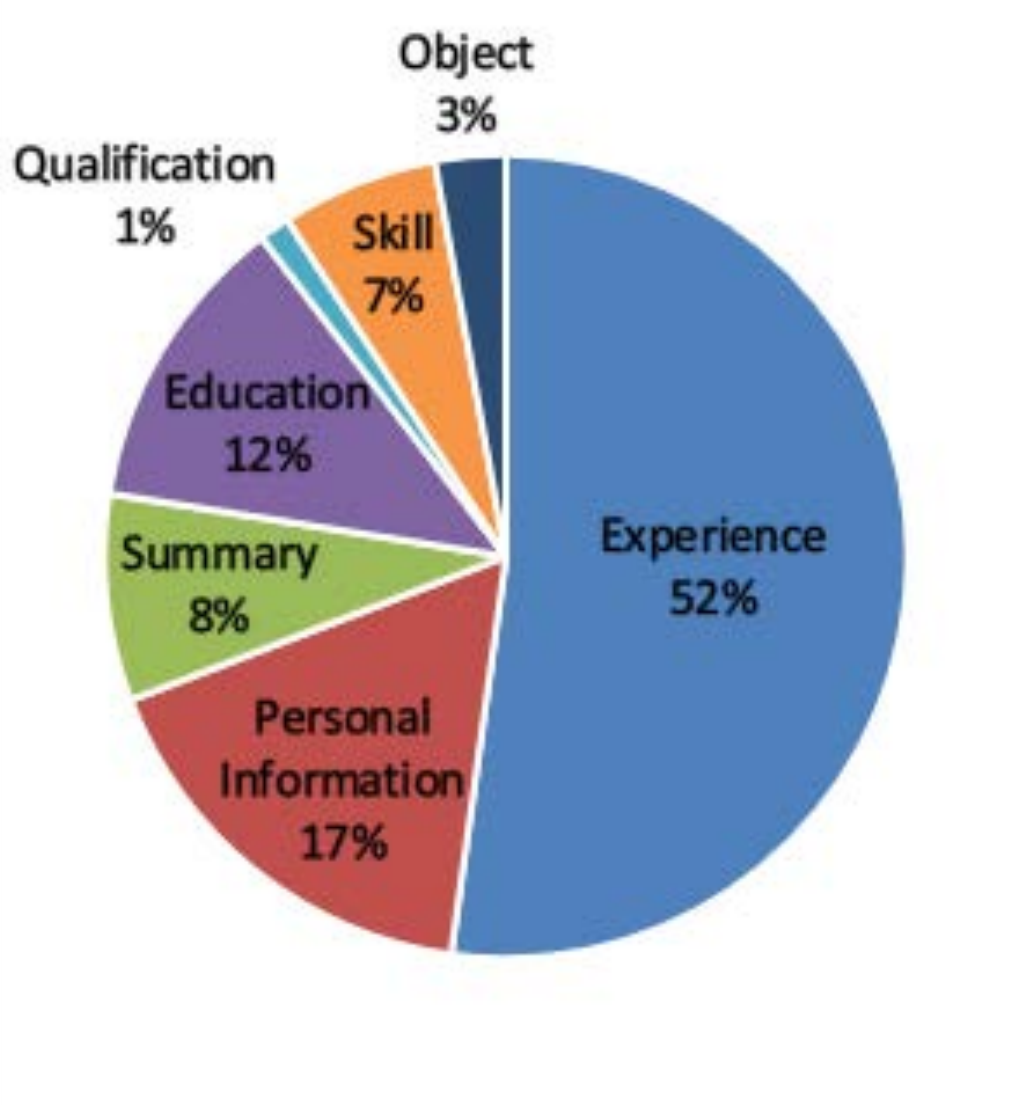}
\caption{Fan chart of the percentage of each category of the resume corpus.}
\label{figure4}
\end{figure}

In the final section, we analyze the sample distribution of the constructed resume corpus.Figure \ref{figure4} shows that the category with the most significant proportion in the resume corpus is \emph{experience}, which accounts for half of the resume text. In addition, the three categories that account for the least in the resume corpus are \emph{skill}, \emph{object}, and \emph{qualification}, which account for only 7$\%$, 3$\%$, and 1$\%$.Conclusively, resume text is a very easy sample imbalance for experimental subjects. Thus, the resume corpus also vigorously tests the model's learning capability for categories with sparse samples in the training dataset. Hence, we plotted the conflation matrix of RoBERTa and T5 models. It is used to analyze the learning ability of the two models for sample-sparse categories in the dataset.
\par
\begin{figure}[!ht]
\centering
\begin{subfigure}{1\linewidth}
\centering
\includegraphics[width=1\linewidth]{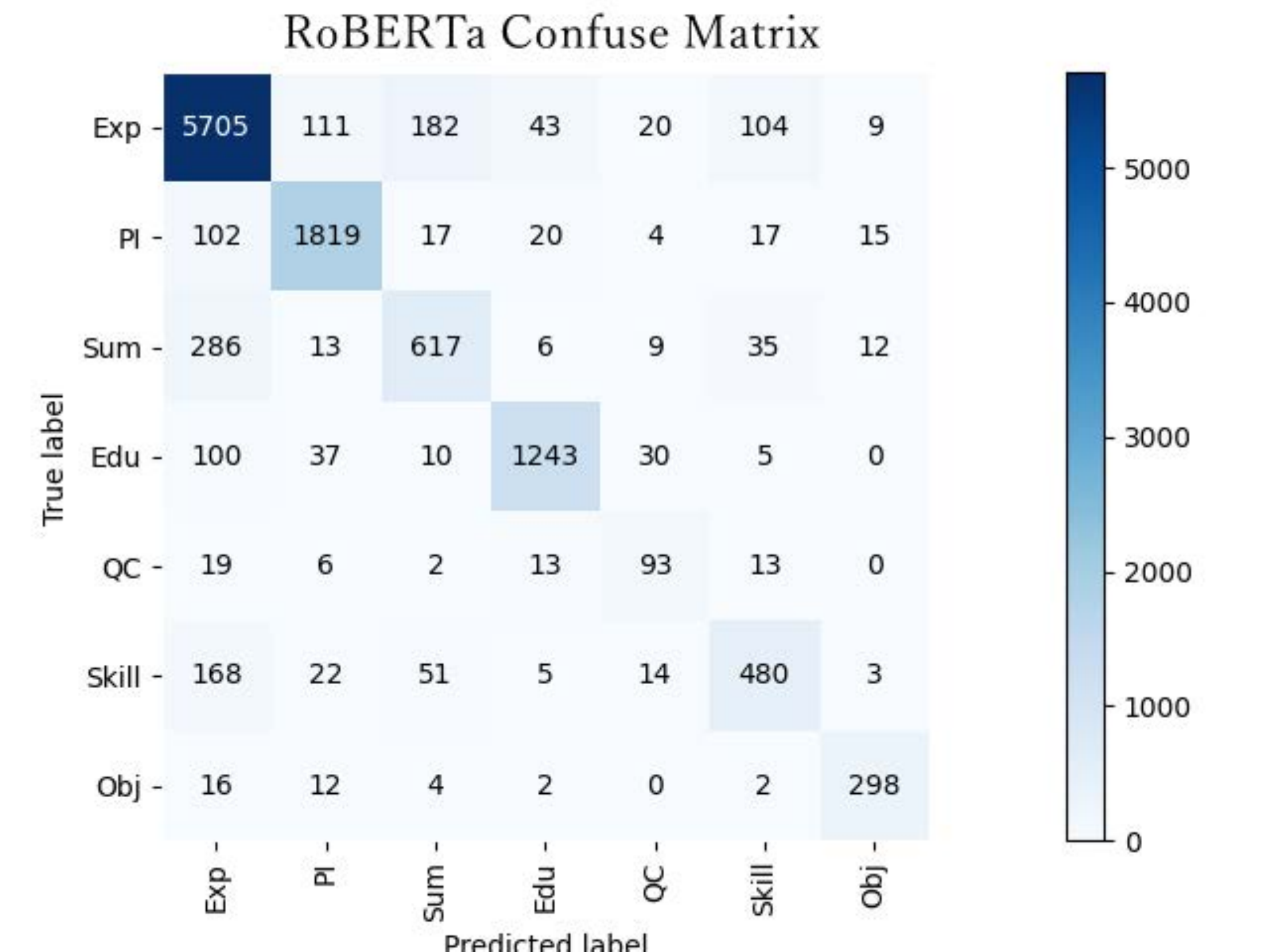}
\label{fig:roberta}
\end{subfigure}

\begin{subfigure}{1\linewidth}
\centering
\includegraphics[width=1\linewidth]{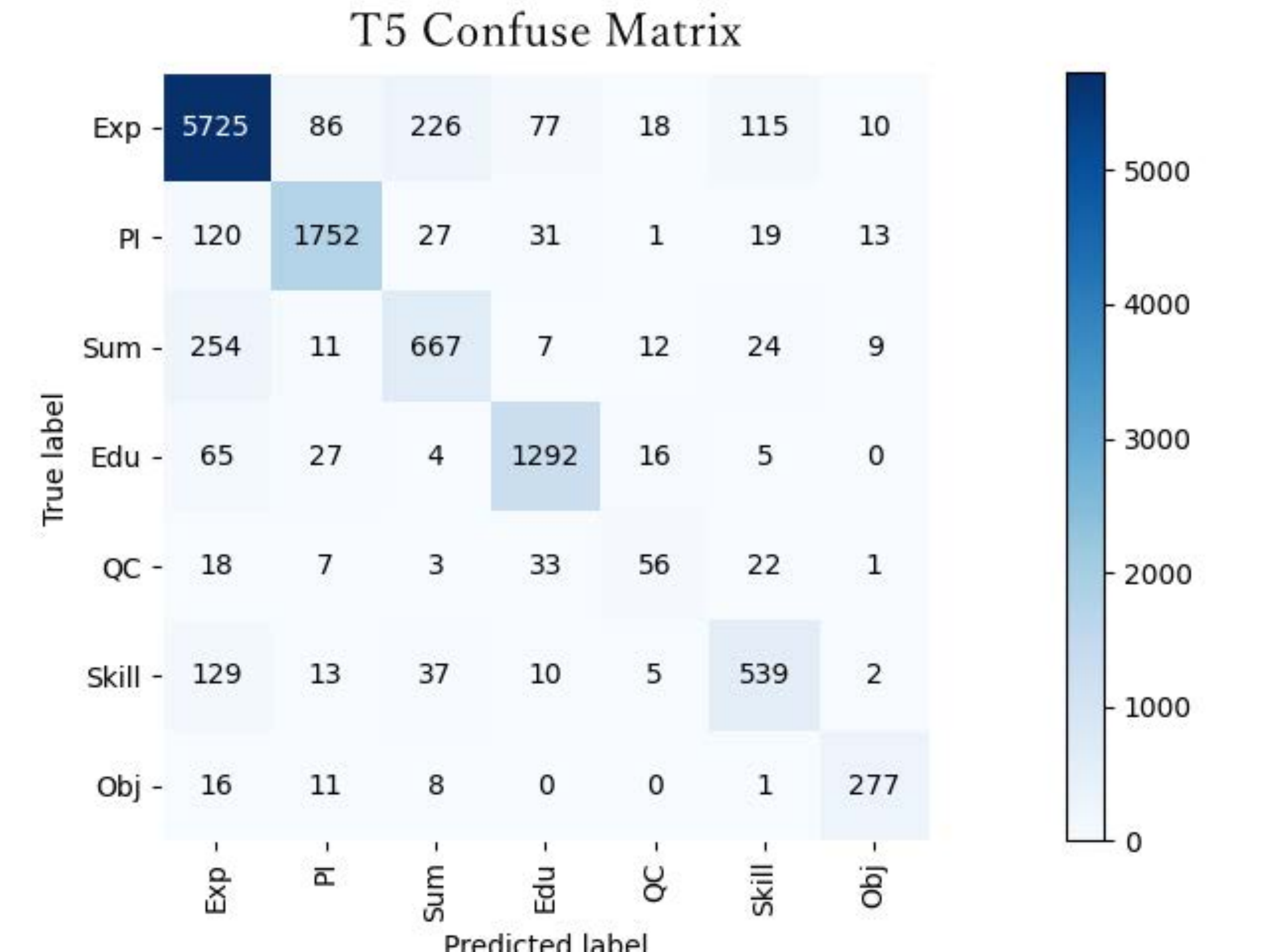}
\label{fig:t5}
\end{subfigure}
\caption{Confuse Matrix of RoBERTa$_{\mbox{\scriptsize large}}$ and T5$_{\mbox{\scriptsize large}}$ model in test set.}
\label{figure5}
\end{figure}

 As shown in the figure \ref{figure5}, we can see the confusion matrix of RoBERTa and T5 models. First,the RoBE-RTa model is better for classifying \emph{qualification} with the least number of samples.
Secondly, the T5 model is slightly better than the RoBERTa model in terms of overall category 
classification results. The above results also demonstrate that our constructed resume corpus is highly unbalanced. However, if the model has strong performance, it can still learn the features of the corresponding category from very few samples.

\section{Conclusion}
In this paper, we improve the classification labels of the original English resume corpus. Furthermore, it doubled the number of samples size. The final tests and analyses also show the reliability of the newly constructed resume corpus. In future work, we will explore how to solve the sample imbalance problem of the resume corpus. Make the model learn effectively even for small sample categories.

\bibliography{custom}
\bibliographystyle{acl_natbib}

\appendix

\section{Appendix A}
\label{sec:appendix}

\begin{figure*}[!b]
\centering
\includegraphics[width=445 pt]{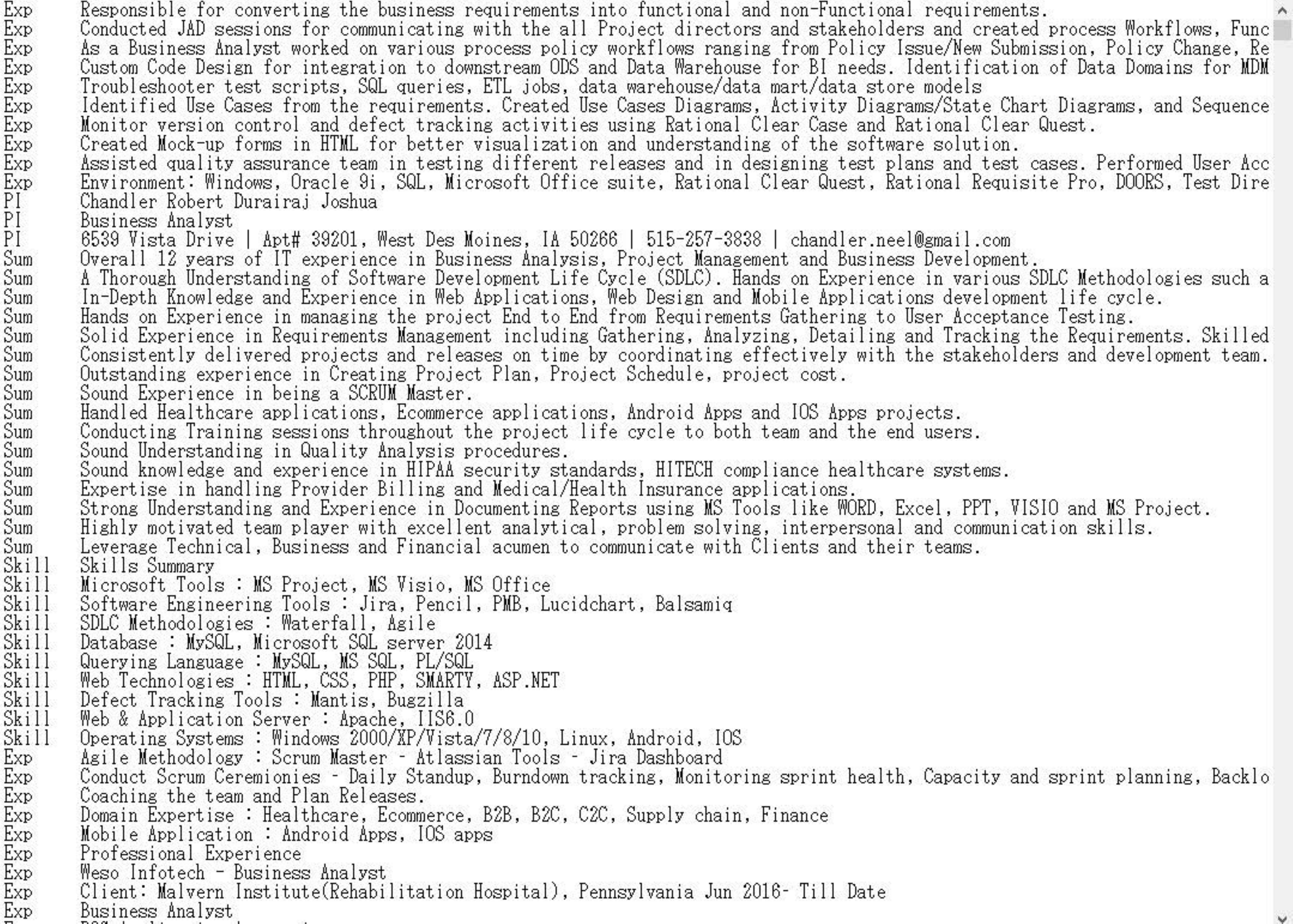}
\end{figure*}

\end{document}